\documentclass[sn-basic,iicol]{sn-jnl}%

\usepackage{graphicx}%
\usepackage{multirow}%
\usepackage{amsmath,amssymb,amsfonts}%
\usepackage{amsthm}%
\usepackage{mathrsfs}%
\usepackage[title]{appendix}%
\usepackage{xcolor}%
\usepackage{textcomp}%
\usepackage{manyfoot}%
\usepackage{booktabs}%
\usepackage{algorithm}%
\usepackage{algorithmicx}%
\usepackage{algpseudocode}%
\usepackage{listings}%

\begin{document}

\title[Article Title]{Learning to Summarize and Answer Questions about a Virtual Robot's Past Actions}


\author*[1]{\fnm{Chad} \sur{DeChant}}\email{chad.dechant@columbia.edu}

\author[2]{\fnm{Iretiayo} \sur{Akinola}}\email{iakinola@nvidia.com}

\author[1]{\fnm{Daniel} \sur{Bauer}}\email{bauer@cs.columbia.edu}

\affil[1]{\orgdiv{Computer Science Department}, \orgname{Columbia University}, \orgaddress{\street{500 West 120 Street} \city{New York}, \postcode{10025}, \state{NY}, \country{United States}}}

\affil[2]{\orgname{NVIDIA}, \orgaddress{\street{4545 Roosevelt Way NE} \city{Seattle}, \postcode{98105}, \state{WA}, \country{United States}}}


\abstract{

When robots perform long action sequences, users will want to easily and reliably find out what they have done. We therefore demonstrate the task of learning to summarize and answer questions about a robot agent's past actions using natural language alone. A single system with a large language model at its core is trained to both summarize and answer questions about action sequences given ego-centric video frames of a virtual robot and a question prompt. To enable training of question answering, we develop a method to automatically generate English-language questions and answers about objects, actions, and the temporal order in which actions occurred during episodes of robot action in the virtual environment.
Training one model to both summarize and answer questions enables zero-shot transfer of representations of objects learned through question answering to improved action summarization. 
}




\maketitle


\section{Introduction}\label{sec1}

As robots become more capable and are entrusted with more tasks, it will be increasingly important to reliably keep track of what they do. However, robots will routinely perform roles that make direct supervision of them difficult or impossible. A robot may, for example, be used to move many loads of construction material from place to place or perform household chores. In both cases, real time human oversight would be impractical. It will therefore be necessary to develop methods to monitor and record the actions of such agents and provide that information at a later time to a human. One way to do that is to develop the capability for robots to summarize and answer questions about their actions in natural language. 

Summaries, rather than complete records, will be particularly useful as action sequences become longer. They will also be challenging to produce because it will be necessary to identify the most important actions and, very often, to describe those actions using higher level abstract terms. Summaries may not fully address everything that a user wants to know about a robot's actions so a user may want to ask questions about what a robot did or saw during a particular action sequence. Providing summaries and answering questions are therefore complementary skills which we would want robotic agents to possess.

\begin{figure*}
\begin{center}
\centerline{\includegraphics[width=\textwidth]{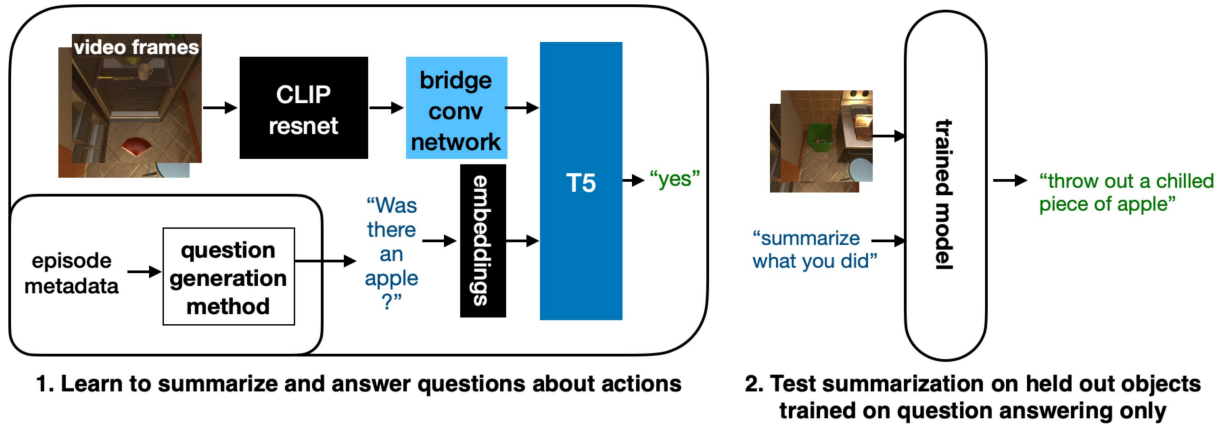}}
    \caption{Visual presentation of model and method for producing zero-shot summaries involving novel objects. Step 1 illustrates the full model: input (at the left) includes video frames as well as episode metadata describing the environment as the agent saw it. The components of the model in black (\textsc{clip} Resnet and the word embeddings) are pretrained and remain frozen during our training process, while the light blue module (the vision-to-T5 bridge network) is trained from scratch. The dark blue module, a pretrained T5, which outputs the final question answer or summary, is fine-tuned during training. Step 2 demonstrates zero-shot summarization using a previously trained model which was not trained to summarize episodes with some of the objects in the newly presented episode.}
\label{banner_diagram}
\end{center}
\end{figure*}

We demonstrate a model that learns such summarization and question answering skills by making use of the capabilities of a large language model (\textsc{llm}). Although much work has gone into training robot agents to follow natural language instructions, little work has addressed reporting back what a robot has done, which might be seen as the flip side of instruction following. Fortunately, existing datasets designed for instruction following tasks can be repurposed and augmented to serve as a training ground for robot action summarization and question answering. We make use of and augment the popular \textsc{alfred} dataset \citep{shridhar2020alfred} which provides ego-centric video frames of episodes of robot action sequences in a virtual environment.

Our main contributions are:\\
\emph{Summarization of actions.} We demonstrate summarization of robotic actions in both short and long summaries from video frames in a multimodal model that incorporates vision and fine-tunes a pretrained T5 \textsc{llm} \citep{raffel2020exploring}.\\
\emph{Answering questions about actions.} The same model is jointly trained to answer questions about robotic actions, including questions about actions performed, objects seen, and the order in which actions were performed.\\
\emph{Zero-shot transfer from question answering to summarization.} 
We show that an \textsc{llm}-based system trained to answer questions about held-out objects can faithfully produce summaries about those objects in a zero-shot manner, even though the objects are not in the summarization task training set. This demonstrates the transfer of representational knowledge from the question answering tasks to the summarization tasks.\\
\emph{Automatic generation of questions and answers.} We develop a method to automatically generate questions and answers using an existing dataset and its associated virtual environment and release a dataset of such questions and answers.

\section{Method}\label{sec2}

Our objective is to generate a summary or question answer in natural language $a \in \mathcal{L}$ of a long horizon robotic task, given the history of observations $o \in \mathcal{O}$ that the robot experienced during the task and a question or summarization prompt $q$.
We define the robot experience/trajectory as $\tau = \{ (o_{0}, ...) \}$.
We seek to learn a function $\mathcal{F}_{\theta}$ such that: $a = \mathcal{F}_{\theta} (\tau, q)$.

\subsection{Repurposed dataset}
Our approach requires egocentric video or video frames, a description of an agent's actions during an episode, and information about the environment the agent operates in, particularly the locations of objects it encounters. For the purposes of the current investigation we use episodes from the \textsc{alfred} dataset. An episode of robot state-action trajectory in the original dataset has four different kinds of representation which we make use of, either as-is or transforming them in some ways. The following list of dataset elements lays out the way they are used in this work as well as noting their original purpose and description in the \textsc{alfred} dataset:\\
1) \emph{Short summaries}: Human-generated natural language one sentence summaries of the whole action sequence (called ``goal descriptions'' in the original dataset).\\
2) \emph{Long summaries}: High level narratives of the robotic agent's actions, provided in the original dataset in the form of action plans in the structured Planning Domain Description Language (\textsc{pddl}) \cite{mcdermott1998pddl}. We convert the terms used in \textsc{pddl} to natural language: for example, ``GotoLocation'' becomes simply ``go to'' and some object names become two English words instead of one word (e.g. ``coffeemachine'' becomes ``coffee machine''). We also break these long summaries up to form questions, as described in the next subsection.\\
3) \emph{Natural language action description sentences}: Natural language step by step descriptions of the actions taken in each episode, written by humans, which were used as instructions in the original dataset. These are used here to form some of the questions, as described in the next section.\\
4) \emph{Video, images, and visual features}: Raw video of a task episode as well as still frames from the video. We use a pre-selected subset of frames in the original dataset, leaving the question of frame selection to future work.

\begin{figure*}
\begin{center}
\centerline{\includegraphics[width=.9\textwidth]{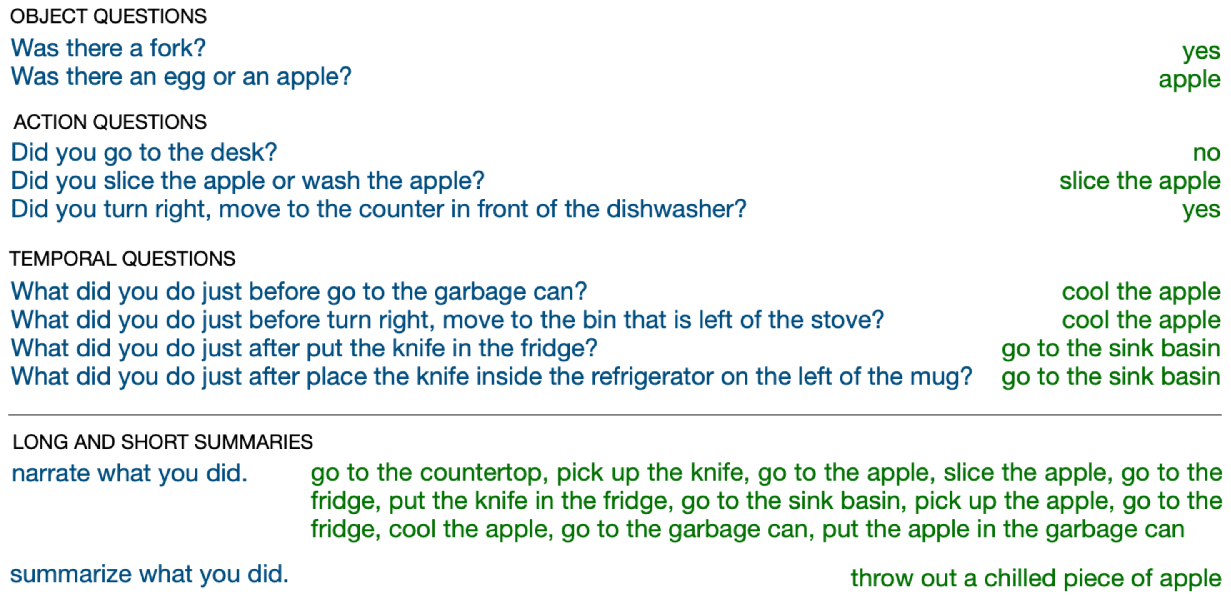}}
    \caption{Sample questions (on the left, in blue) and expected answers (on the right, in green), broken up into question type, along with the prompts for long and short summaries, at the bottom.}
\label{examples_text}
\end{center}
\vspace{-2.5mm}
\end{figure*}

\subsection{Automatic generation of questions and answers}
We develop a Q\&A generation algorithm that produces questions and answers about episodes of robots interacting with an environment. After initial pre-processing, the algorithm can be used in a partly online fashion during training or as a one-time off-line dataset generation step which produces a set of static questions and answers. We train models in an online fashion and provide performance metrics from the static validation sets of questions and answers we release with this work.

In addition to the elements already present in the original dataset enumerated in the previous subsection, we use the \textsc{ai2thor} environment \citep{kolve2017ai2} to rerun the agent trajectories for each episode in the dataset and capture metadata present while the agent is in the environment. This metadata is used to generate questions and includes information about objects encountered in the virtual environment and the order in which the robot sees and interacts with them. Though here we use one particular existing dataset and environment, our approach is general and can be used in other cases where similar data can be captured.

The algorithm produces nine types of questions in three broad categories (see Figure \ref{examples_text} for examples of each type):

(1) \textbf{Object questions} about the presence of objects in the environment, both those the agent interacted with and those it only saw. There are two kinds of object question: ``object yes/no'' questions of the form, ``was there an $<$object$>$?", which require only ``yes'' or ``no'' answers and ``object either/or'' questions of the form, ``was there an $<$object A$>$ or $<$object B$>$?'' which require the model to output the name of the object present. Our algorithm uses the metadata of all objects visible in the environment to ensure that only one of the objects in an either/or question will have been seen during an episode. The algorithm samples objects with negative answers in proportion to their appearance in the dataset so that the model cannot, for example, learn to always answer, ``no'', for seldom-seen objects. Questions with ``yes'' and ``no'' answers are presented with equal frequency.

(2) \textbf{Action questions}, which ask about actions the agent performed. The two types of question --- ``action yes/no'' and ``action either/or'' --- follow the structure of the respective object questions explained above. There are two subtypes of the ``action yes/no'' questions: ``simple action yes/no'' uses the relatively simple language converted from \textsc{pddl} for both the questions and answers. ``Complex action yes/no'' uses the raw human-generated description of each action step to pose the ``yes/no'' question. ``Action either/or'' questions present an either/or choice between two actions described in the simpler language of the converted \textsc{pddl} plans.

(3) \textbf{Temporal questions} about the order in which actions were performed, of two primary kinds. The first kind --- ``just before'' questions --- asks what action was performed immediately before a named action (``what did you do just before $<$action description$>$?") while the second --- ``just after'' questions --- asks what action was performed immediately following the named action (``what did you do just after $<$action description$>$?"). If an action occurs more than once in an episode it will not appear in a temporal question to avoid ambiguity.

Each of these types of temporal questions has two subtypes. The first is asked using the simpler description of actions from converted \textsc{pddl} (e.g. the first temporal question in Figure \ref{examples_text}) while the second uses a human-generated action description sentence to formulate the question (e.g. the second temporal question in Figure \ref{examples_text}) . These descriptions are longer, contain more diverse word choice, and sometimes mention irrelevant details. The answers to both question subtypes are in the simpler action description format. We suggest that this distinction between enabling the model to answer both simple and more complexly-worded questions while only answering in simpler language is desirable because while a robot agent should be able understand questions phrased in a variety of ways, for the sake of clarity such an agent should not produce similarly varied answers, but instead generate only simple, consistent language. 

In addition to these questions and answers, we also prompt the model to produce two kinds of summaries:\\
(1) \textbf{Short summaries} are the short one sentence descriptions of the action sequences written by human annotators as provided in the original dataset. We train the model to output a summary of a given episode with the text prompt, ``summarize what you did."\\
(2) \textbf{Long summaries}, which are the longer narratives of actions converted from \textsc{pddl} to natural English. Although these are meaningfully longer than the one sentence summaries, they are significantly shorter than a step by step account of every low level action the virtual robot performed (e.g. move ahead, turn, look up, etc.). The model is trained to output a long summary of an episode with the prompt, ``narrate what you did.''

\subsubsection{Dataset of questions and answers}
We will release both the code to generate the questions and answers as well as a static set of premade questions and answers aligned to episodes in the \textsc{alfred} dataset. The static dataset was generated to produce up to ten question tokens per question type for each episode; in some cases there are fewer than ten such question tokens per episode because not all question types can produce ten question tokens for a given episode.

The entire static question and answer set contains 486,704 questions paired to episodes in the \textsc{alfred} dataset's training set, 18,891 questions paired to its seen environments validation set, and 19,097 in its unseen environments validation set.

\subsection{Joint summarization and question answering model}
We present a learned algorithm that takes as input ego-centric video frames of a virtual mobile robot along with a natural language question or summarization prompt and produces an answer or summary in response.

Our full neural network model (see the breakdown on the left in Figure 1) combines several components. Video frames are fed into a frozen Resnet network \citep{he2016deep} pretrained as part of the \textsc{clip} model \citep{radford2021learning}. We extract the output of the last convolutional layer and feed it into a three layer convolutional network trained from scratch, which acts a bridge network between the Resnet and the next step in the pipeline: a pretrained T5 transformer \textsc{llm} \citep{raffel2020exploring} (``t5-base''in the Hugging Face library \citep{wolf2020transformers}) which we fine tune. While the T5 model was pretrained only on language data, we use it for simultaneous language and visual input, following other work which has shown the ability of language model transformers to process multimodal data \citep{lu2022frozen, tsimpoukelli2021multimodal}.

\begin{table*}
\centering
\begin{tabular}{lcc|cc}
\hline
\textbf{Question / prompt}   & \textbf{Seen envs} &  &   \textbf{Unseen envs} & \\
\hline

 & \textbf{Accuracy} & \textbf{Precision} & 
\textbf{Accuracy} & \textbf{Precision} \\
\hline

Object yes/no & .954 $\pm$ .007 & - &   .907 $\pm$ .010 & - \\
Object either/or & .990 $\pm$ .003 & .990 $\pm$ .003 & .966 $\pm$ .009 & .968 $\pm$ .010\\
\hline
Simple action yes/no & .975 $\pm$ .001 & - &.892 $\pm$ .004 & - \\
Complex action yes/no & .935 $\pm$ .003 & - & .895 $\pm$ .004 & -\\
Simple action either/or & .988 $\pm$ .003 & .995 $\pm$ .001 & .923 $\pm$ .019 & .963 $\pm$ .009\\

\hline
Simple action just before & .948 $\pm$ .004 & 976 $\pm$ .003 & .865 $\pm$ .012 & .957 $\pm$ .004 \\
Complex action just before & .927 $\pm$ .007 & .967 $\pm$ .004  & .818 $\pm$ .013 & .939 $\pm$ .005\\
Simple action just after & .959 $\pm$ .002 & .983 $\pm$ .002 & .815 $\pm$ .017 & .936 $\pm$ .004 \\
Complex action just after & .911 $\pm$ .009 & .952 $\pm$ .004  & .730 $\pm$ .015 & .887 $\pm$ .002\\

\hline
Long Summary  & .850 $\pm$ .005 & .969 $\pm$ .011 & .475 $\pm$ .035 & .945 $\pm$ .004\\
\hline
 & \textbf{\textsc{rouge}} & \textbf{\textsc{bleu}} & 
\textbf{\textsc{rouge}} & \textbf{\textsc{bleu}} \\
\hline

Short Summary  & .571 $\pm$ .000 & .556 $\pm$ .006 & .517 $\pm$ .004 & .504 $\pm$ .022\\
Long Summary  & .981 $\pm$ .000 & .969 $\pm$ .000 & .922 $\pm$ .004 & .880 $\pm$ .010\\

\hline

\end{tabular}
\caption{\label{tab:metrics}
Accuracy and precision scores for question and summary outputs by output type, including standard deviation. \textsc{rouge} and \textsc{bleu} scores also given for summaries. Results shown are from two validation sets: those based on episodes in virtual environments seen during training are on the left, unseen environments on the right. None of the actual episodes themselves, of either  type, are found in the training set. Precision scores are not shown for ``yes/no''answers where such scores must equal the accuracy scores. Results are averaged from three models with different random seeds, all tested on the set of static held-out questions.}
\end{table*}

The tokens of the natural language questions and summarization prompts are embedded using the T5 model's pretrained embeddings. We concatenate these text embeddings with the image vector representations yielded by the bridge network. As T5 is an encoder-decoder model it is able to generate encoded representations of the images conditioned on the given question or prompt. We train a single model to answer all questions and produce long and short summaries so that it must learn to generate representations useful for all of these tasks. During an epoch of training we iterate through each episode in random order. For each episode, the model must produce long and short summaries and answer one question of each of the nine question types (when such a question exists for that episode).

\subsection{Zero-shot summarization after question answering}
We are interested in the possible interaction between question answering and summarization abilities within the model, in particular if representations of objects transfer between these tasks. We therefore alter the training regime to leave some objects out of the summarization training set and measure whether the model is still able to produce accurate summaries about interactions with the objects. In these experiments, we first randomly select a set of five objects from among the most common thirty objects in the dataset (excluding the top ten). We then identify all episodes whose long summaries contain those objects (i.e. any episode in which the virtual robot interacts with those objects) and set them aside as a `held-out' set. The model is then trained on questions and answers involving all episodes, including the held-out episodes, but is not trained to produce either long or short summaries of the held-out episodes.

\section{Results}\label{sec3}

\subsection{Summarization and question answering}
We find that our model performs very well on both short and long summarization tasks and on the questions from our Q\&A generation algorithm. 
Table \ref{tab:metrics} presents results for all question and summarization types. An answer is considered accurate if it completely matches the target answer. \textsc{bleu} \citep{papineni2002bleu} and \textsc{rouge} \citep{lin2004rouge} scores are also given for the two summary types. The \textsc{bleu} score is a measure of how well the generated text matches the ground truth text, penalizing words and phrases which are not present in the ground truth while \textsc{rouge} measures how much of the ground truth text is present in the generated text, penalizing words and phrases which are missing from the generated text. Unigram precision scores measure the percentage of generated words which are in the ground truth text and are given for question answering tasks which require more than one word as an answer. As the short summaries are more lexically diverse, binary accuracy measures are less appropriate so precision scores are given for the short summaries. 

A few patterns in the results can be seen.
%
%
First, the performance generally varies depending on how much generated text must be produced in an answer. Longer answers provide more opportunities for errors so performance when measured by the strict metric of complete accuracy tends to be worse. This is particularly true for the question which asks for a long summary of the agent's action, which has the worst results according to the all-or-nothing accuracy metric. 

\begin{table}
\begin{tabular}{lc}
\hline
\textbf{Question}   & \textbf{Error Overlap} \\
\hline

Object yes/no & .054 $\pm$ .023 \\
Object either/or & .061 $\pm$ .029\\
\hline
Simple action yes/no & .370 $\pm$ .059 \\
Complex action yes/no & .222 $\pm$ .051 \\
Simple action either/or & .464 $\pm$ .058\\

\hline
Simple action just before & .441 $\pm$ .071 \\
Complex action just before & .414 $\pm$ .028 \\
Simple action just after & .725 $\pm$ .020  \\
Complex action just after & .414 $\pm$ .028 \\

\end{tabular}
\caption{\label{tab:overlap_metrics}
Overlap of missing objects between questions and long summaries by question type, averaged over three models tested on the static held out valid unseen set. Overlap here is the number of missing word errors per question type for which the long summaries are also missing the same word in the same episode, as a percentage of all missing word errors per question type.}
\end{table}
Second, ``either/or'' questions have better accuracy than their corresponding ``yes/no'' questions. This could be because asking if, for example, an action was performed is made easier when it is a choice between two actions so that any uncertainty the model has about one of the actions may be offset by its certainty about the other option. It is also possible that the model has a harder time connecting the meaning of the ``yes/no'' answers back to the input, particularly since most of the questions require outputting an object or action name, not just a ``yes/no''.

Third, it might be expected that questions about the order that actions took place would be significantly more difficult for the model to interpret than those about the mere occurrence of those actions. Surprisingly, we find that in most cases the model's performance on temporal questions is very similar to that on the other questions.

\begin{table*}
\centering
\begin{tabular}{lccc}
\hline

\textbf{Model trained on} & \textbf{seen envs} & 
\textbf{unseen envs} \\
\hline

All questions & .519 $\pm$ .067 & .302 $\pm$ .156 & \\
Temporal questions & .566 $\pm$ .118 & .299 $\pm$ .085 &\\
All other non-temporal questions & .006 $\pm$ .007 &.000 $\pm$ .000 & \\
\hline

No Q\&A training on held out objects & .000 $\pm$ .000 & .000 $\pm$ .000 \\
\hline
All objects and questions - nothing held out &  .850 $\pm$ .005 & .475 $\pm$ .035 \\

\end{tabular}
\caption{\label{tab:zeroshot_accs}
Accuracy of zero-shot long summarization when transferring representations learned from question answering to producing long summaries, broken down by question type used to learn the objects held out from summarization training. Results shown for episodes containing held-out objects in the validation sets in seen and unseen environments. The bottom two rows show a baseline with no question answering training on the held-out objects --- and therefore no transfer --- and a comparison to the fully trained model with nothing held out.}
\end{table*}

\begin{figure*}
\begin{center}
\centerline{\includegraphics[width=\textwidth]{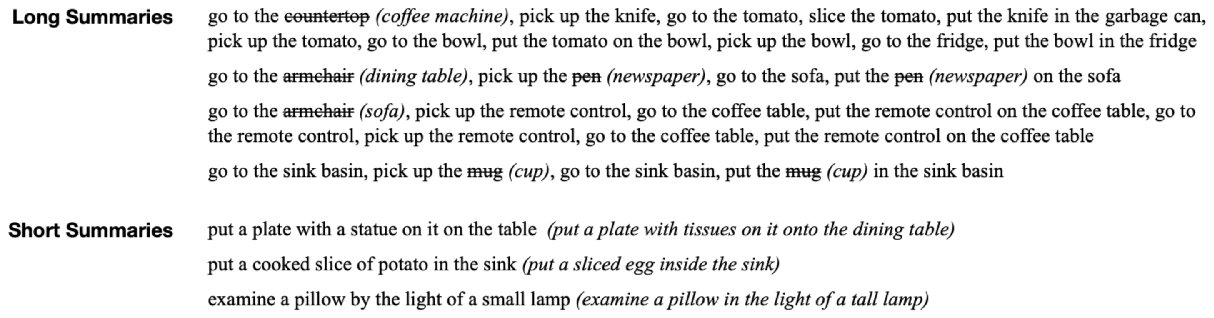}}
    \caption{Example errors in generated long and short summaries. Errors in the long summaries are indicated with strikethrough text (with the correct text following in italics and parentheses). Generated short summaries appear to the left of the correct summaries, which are in italics.}
\label{errors}
\end{center}
\vspace{-3.5mm}

\end{figure*}

The model tends to make two kinds of errors when generating anything other than ``yes/no'' answers. It sometimes misidentifies objects, especially small ones, and particularly in the unseen environments. It also sometimes uses a different description for a location than the ground truth annotation, in some cases doing so in a way that is nevertheless consistent with the action as seen in the episode. For example, the ground truth annotation may read, ``go to the apple'' while the model outputs, ``go to the counter'' when the apple is on the counter. See Figure \ref{errors} for examples of errors in short and long summaries generated by the model.

The errors made by the model display some consistency between the different questions asked and between the questions and summaries. For example, in one episode of the validation seen set which involves moving a book, it consistently mistakes the book for a pen, answering a ``just before'' question with, "put the pen on the desk," producing a short summary, ``put two pens on the right side of the desk,'' and beginning the long summary with, ``go to the side table, pick up the pen...''. There is a marked difference in the consistency of these errors depending on question type, however, as we show in Table \ref{tab:overlap_metrics}. We measure this consistency by counting what fraction of particular objects omitted from the model's answers to a given question type is also missing from the corresponding long summaries about that episode. This fraction is compared for different question types. We find that questions which require generating both an action and an object together have the highest degree of overlap in which objects they fail to identify and which are also missing in the long summaries; the temporal ``just before / just after'' answers in particular show high consistency with the long summaries. We hypothesize that the representations which the model uses for summarization align better with those it uses for the question types where there is higher overlap of missing words. 

\subsection{Zero-shot summarization via question answering}
Can question answering improve the ability to summarize? We find that when the model is trained to answer questions about episodes involving all objects, it is then able to go on to summarize episodes with objects which it has not been trained to include in summaries. Table \ref{tab:zeroshot_accs} displays a breakdown of zero-shot performance on long summaries. For comparison, results when nothing is held out --- the standard case detailed in table 1 --- and for a model not trained to answer questions on the held out set are included. These comparisons show that while zero-shot summarization is not as accurate as fully supervised summarization, training on the auxilliary question-answering task is significantly better than not. A model not trained to answer questions on episodes with held out objects is unable to correctly summarize episodes involving those held out objects. It is simply not able to output any of the held out objects' names without having at least seen them during question answering. Training the model to learn to answer questions about the objects through an auxilliary question-answering task leads to clear improvement on the summarization task.

This result suggests that the model is learning representations of objects, or actions involving objects, while learning to answer questions which it can then use when producing summaries. There must be at least some transfer of representational knowledge between the question answering and the summarization tasks within the model. 

Clear improvement with transfer compared to without transfer is also demonstrated in \textsc{bleu} and \textsc{rouge} scores of both short and long summaries in seen and unseen environments (in only one case is there not improvement); see Table 5 in Appendix \ref{sec:app_zero_shot_transfer} for details.

\subsubsection{Impact of question type on zero-shot transfer to summarization}
We have seen that transfer from question answering to summarization occurs. But which questions are most important? In order to further investigate the sharing of representations between question answering and summarization, we rerun the experiments using the same held out protocol, but using focused sets of particular question types. Testing each question type separately allows us to measure whether all questions are equally useful for promoting transfer to summarization. Interestingly, we find that not all questions are equally useful: only the temporal ``just before'' and ``just after'' questions --- which ask what action was performed just before or after a given action --- exhibit the transfer between tasks (see Table \ref{tab:zeroshot_accs} for accuracy metrics on temporal and non-temporal questions). This is true of both subtypes of these questions, i.e. both the simple and complex language versions. On their own, the ``yes/no'' and ``either/or'' questions about objects or actions do not lead to the same zero-shot summarization ability. It is worth recalling here that the answers to the temporal questions were also found to be especially consistent with the long summaries in the missing object errors they contained, which would also suggest a particularly aligned representational space between those tasks (see Table \ref{tab:overlap_metrics}).

We also tested the transfer ability of a model trained in a similar manner but which excluded episodes based on the action verbs they contained rather than the objects. For these experiments, only one action verb at a time and the episodes which contained it were identified as held out items. In none of these cases was the model able to transfer the use of the verb to summaries of the held out episodes. This could be due to the smaller number of actions in the dataset than objects.

\section{Related Work}\label{sec4}

\textbf{RoboNLP} \citet{tangiuchi2019survey} and  \citet{tellex2020robots} offer thorough reviews of language use in the context of robotics. Detailed descriptions of actions such as robots playing soccer \citet{mooney2008learning} or automated driving \citet{barrett2015robot, barrett2017driving} have been generated. These have not involved learning how to report and condense a series of actions into anything like a summary, however. \citet{dechant2021toward} propose robot action summarization as a research direction, suggesting a set of tasks to pursue.

\textbf{Instruction Following} Our proposal is closely related to learning to follow natural language instructions, which has long generated a great deal of interest at the intersection of robotics and natural language processing \citep{winograd1972understanding, dzifcak2009and}. \citet{shridhar2021cliport} train a robotic arm in a virtual environment to perform a range of tasks following natural language instructions and transfer the learned model to a real world robot. \citet{calvin21} introduce a benchmark for so-called long horizon (many step) robotic manipulation tasks following natural language instructions.

Rich simulated environments for language-guided navigation tasks have been introduced in recent years. \citet{anderson2018vision} introduced the Room to Room vision and language navigation dataset, which became the basis for much work in this area. Some of that work has involved learning to generate natural language descriptions of navigation trajectories as a training signal or tool: \citet{nguyen2021interactive} provide feedback to an agent in the Room to Room environment by describing in natural language the paths the agent actually takes so it can learn to compare that to the path it should have taken; \citet{fried2018speaker} learn to generate instructions to augment training data and then, at test time, to evaluate the similarity of routes it might take with the description of the desired route. 

\textbf{Q\&A in robotics} Learning to ask questions has also been worked on as a way for a robotic agent to ask for help or clarification while performing a task \citep{tellex2014asking, thomason2019improving}. \citet{yoshino2021caption} use natural language questions to clarify aspects of how a simple action was performed in response to a question. \citet{datta2022episodic} introduce a form of question answering where the questions are in natural language but the answers take the form of visual highlights of a map to indicate locations. \citet{carta2022eager} propose filling in the blanks within structured language instructions as an auxiliary task for reinforcement learning agents in a 2-D grid world. \citet{gao2021env} introduce a similar Q\&A task in a virtual environment, though without summarization; a slightly different embodied Q\&A task, requiring an agent to seek out answers to questions, is proposed by \citet{gordon2018iqa}.

\textbf{Summarization} There is an extensive body of work on natural language summarization, providing examples and resources for the new task of robot action summarization (see \citet{nenkova2012survey} and \citet{gambhir2017recent} for reviews).

\textbf{Video understanding} Work on understanding video is relevant to our work since we are interested in using video or selected images from video as one of the inputs to summarizing a robot's action in natural language. The task of 'video summarization' in the computer vision community refers to selecting important frames of a video that can, together, serve as a visual summary of the whole video; see \citet{apostolidis2021video} for a review of such techniques. Some work has been done on multimodal summarization from video and text transcripts to natural language summaries; \citet{Palaskar2019MultimodalAS} is one example, going from video and text in the How2 video dataset \citep{Sanabria2018How2AL} to summaries. \citet{Barmann_2022_CVPR} assemble a large question answering dataset for real world video of humans performing actions, requiring significant effort to annotate.

\section{Conclusion}\label{sec5}
We develop a model that can be jointly trained to summarize and answer questions about a virtual robot's past actions. We find that the model learns a representation space which is shared across at least some of the question types and summaries, leading to zero-shot summarization abilities.

This work helps begin a line of research on robot action summarization and question answering. It is important that robots operating in the real world be well supervised by humans and that their actions be understandable.
We suggest that establishing a basic narrative of \emph{what} an agent does is in some ways a prerequisite to further understanding \emph{why} it does something.
Once answering questions about and summarizing robot actions can be performed reliably, we also expect these capabilities to be useful in a variety of other ways, including in training robots and in lifelong learning settings in which robots might receive feedback to the summaries they generate.

It has been recognized for some time that grounding the use of language in machines to the real world is essential for creating \textsc{ai} systems that can actually understand the language they process \citep{harnad1990symbol, bisk2020experience, chandu2021grounding, mcclelland2020placing}. Summarization and question answering about past robot actions provide new avenues for tackling the grounding problem. 

Though this work took place in simulation, the summarization and question answering tasks are not specific to aspects of this or any simulated environment. Future work will explore the application of these tasks to real world robots.

\textbf{Acknowledgements} C.D. was supported by a grant from the Long Term Future Fund.

\textbf{Author contributions} C.D. led the project, performed experiments, and wrote the initial draft of the manuscript. I.A. and D.B. provided supervision, research guidance, and edited and wrote portions of the manuscript.

\backmatter

\section*{Declarations}

\textbf{Funding} C.D. was supported by a grant from the Long Term Future Fund

\noindent
\textbf{Conflict of interest/Competing interests} the authors declare no conflict of interest

\noindent
\textbf{Ethical statements} The work performed for this paper made no use of human subjects. 

\noindent
\textbf{Code availability} Code and data will be released upon publication

\begin{appendices}

\section{Neural network model and training details}\label{secA1}

\begin{table*}
\centering
\begin{tabular}{lcc|cc}
\hline
\textbf{Question / prompt}   & \textbf{Seen envs} &  &   \textbf{Unseen envs} & \\
\hline

 & \textbf{Accuracy} & \textbf{Precision} & 
\textbf{Accuracy} & \textbf{Precision} \\
\hline

Object yes/no & .629 $\pm$ .034 & - &   .636 $\pm$ .030 & - \\
Object either/or & .632 $\pm$ .010 & .636 $\pm$ .034 & .635 $\pm$ .008 & .640 $\pm$ .012\\
\hline
Simple action yes/no & .503 $\pm$ .029 & - &.533 $\pm$ .038 & - \\
Complex action yes/no & .486 $\pm$ .010 & - & .508 $\pm$ .032 & -\\
Simple action either/or & .526 $\pm$ .028 & .710 $\pm$ .001 & .535 $\pm$ .017 & .725 $\pm$ .015\\

\hline
Simple action just before & .479 $\pm$ .016 & .752 $\pm$ .005 & .517 $\pm$ .007 & .775 $\pm$ .014 \\
Complex action just before & .539 $\pm$ .007 & .766 $\pm$ .004  & .613 $\pm$ .025 & .826 $\pm$ .017\\
Simple action just after & .315 $\pm$ .022 & .706 $\pm$ .017 & .349 $\pm$ .039 & .728 $\pm$ .021 \\
Complex action just after & .256 $\pm$ .018 & .658 $\pm$ .003  & .325 $\pm$ .019 & .695 $\pm$ .003\\

\hline
Long Summary  & .005 $\pm$ .002 & .492 $\pm$ .094 & .000 $\pm$ .000 & .471 $\pm$ .101\\
\hline
 & \textbf{Rouge} & \textbf{Precision} & 
\textbf{Rouge} & \textbf{Precision} \\
\hline

Short Summary  & .264 $\pm$ .027 & .496 $\pm$ .058 & .263 $\pm$ .022 & .485 $\pm$ .059\\
\hline

\end{tabular}
\caption{\label{tab:ablation}
Ablation of video frames baseline: results for a model trained to answer questions and produce summaries when trained with questions and answers as usual but with each question and answer pair and summarization task paired to identical visual input (i.e. each episode's observations are replaced by a single, static set of observations that do not vary from episode to episode), thereby completely depriving the model of any useful visual information with which to answer the question.}
\end{table*}

\subsection{Neural network}

We train a bridge network to downsize and link the output of the last convolutional layer of a pretrained \textsc{clip} Resnet-50 network \citep{radford2021learning} with a pretrained T5 transformer (``t5-base'') from the Hugging Face library \citep{raffel2020exploring, wolf2020transformers}. 

\textbf{Bridge network architecture}

Input: 2048 x 7 x 7 Resnet features

Conv layer 1 (2048, 1024, 1)

Conv layer 2 (1024, 128, 1)

Conv layer 3 (128, 32, 1)

Fully connected layer (1568, 768)

\textbf{Total number of parameters}

The \textsc{clip} Resnet-50 network has a total of 102,007,137 parameters.

The bridge network described above has a total of 3440,864 parameters.

The T5 Transformer network we fine-tune has a total of 222,903,552 parameters.

The bridge network and T5 networks together (the complete model we train / fine-tune) have a total of 226,344,416 parameters.

\subsection{Training information}

We primarily used two  Titan Xp and three \textsc{nvidia} A6000 \textsc{gpus}. When training one of our models with all questions and training data, a Titan Xp took approximately 3 days and an A6000 approximately 1.5 days to train for 100 epochs. Though we did not track it, a rough estimate of total \textsc{gpu} time during initial exploration of this problem and the work reported here is 5000 hours.

Hyperparmeters we tested variations of include the optimizer (Adam, AdamW, AdaFactor, AdamW --- Adam was used in all experiments reported here); the learning rate (.001 was used as the initial learning rate in all experiments reported here); and network architecture choices for the bridge network which connected the \textsc{clip} Resnet convolutional layer outputs and the T5 transformer (layer sizes, number of layers, batchnorm, dropout). The random seed was not one of the hyperparameters tuned; we used three random seeds to produce all of the results in the paper, which are averaged across three runs with different random seeds and, in the case of the experiments involving held out objects, three randomly chosen sets of five objects.

We use the dataset's valid seen set as our validation set with which to choose hyperparameters and select the epoch for results to report. We use the accuracy metric of the long summarization task as the measure to select the best epoch. We then report the results of short summarization and all questions in that epoch. These are not the best epochs reached for each of the questions but we report the results from a single epoch to be consistent. The epoch we report best results from the valid seen set is also rarely the best epoch for the valid unseen set but we report the results for the valid unseen set from the same epoch.

Our automatic precision, \textsc{bleu}, and \textsc{rouge} metric was generated from an implementation available through the Hugging Face library.

The \textsc{alfred} dataset was released under an \textsc{mit} License and we release our questions and answers dataset under the same license.

\section{Ablation of visual input}
To investigate how the model would perform on summarization and question answering tasks in the absence of any meaningful input, we trained the model as usual but instead of presenting a sequence of images corresponding to each episode, we presented only one sequence of images from one episode for every summarization prompt and question. Because the T5 language model does not initially have any ability to represent visual information and only learns that through the course of the multimodal training process, it is very unlikely that the model trained with this ablation of meaningful visual inputs could learn anything about visual inputs. The output on validation set summarization and question answering prompts would therefore only be a reflection of what the model has learned about the regularities in the text portion of the dataset, e.g. what actions are more likely to follow from other actions, regardless of the episode visual data. Table \ref{tab:ablation} presents the results of the ablation study on the seen and unseen validation set environments.

Some of the binary questions have accuracies very close to 50\% --- these include the ``simple action yes/no'', ``complex action yes/no'', and ``simple action yes/no''. The ``object yes/no'' and ``object either/or'' questions, on the other hand, have slightly higher accuracies, approximately 63\%, suggesting that model has learned some patterns in the distribution of objects in the dataset. Similarly, the temporal ``just before'' and ``just after'' questions have higher accuracies than a uniformly random choice among possible actions would demonstrate. Our actual model achieves much higher accuracy on all of these tasks, however, demonstrating that it has learned much more than merely the regularities in the dataset.

We initially tested an additional form of question, a question that asked if <Action A> happened before <Action B> in a given episode (note that this type of question differs from the temporal questions included in this work, which ask what action happened \textit{immediately} before or after a given action, not whether an action happened at any point before a given action). We had to exclude this form of question, however, because the model was able to achieve over 80\% accuracy on validation set episodes under the ablated visual input regime. This question was apparently simply too easy given the regularities in the dataset.

\begin{table*}
\centering
\begin{tabular}{lcc|cc}
\hline
\textbf{Question / prompt}   & \textbf{Seen envs} &  &   \textbf{Unseen envs} & \\
\hline

 & \textbf{\textsc{rouge}} & \textbf{\textsc{bleu}} & 
\textbf{\textsc{rouge}} & \textbf{\textsc{bleu}} \\
\hline
\hline

 \textbf{Long Summaries} &  &  & & \\
\hline

All questions & .919 $\pm$ .007 & .888  $\pm$ .018 & .866 $\pm$ .006 & .813 $\pm$ .068 \\
Temporal questions & .940 $\pm$ .004 & .910 $\pm$ .035 & .858 $\pm$ .005 & .798 $\pm$ .004\\
All other (i.e. non-temporal) questions & .810 $\pm$ .003 & .697 $\pm$ .026 & .765 $\pm$ .008 & .651 $\pm$ .088\\
\hline
No Q\&A training on held out objects & .802 $\pm$ .004 & .688 $\pm$ .022 & .806 $\pm$ .003 & .703 $\pm$ .022\\
\hline
All objects and questions - nothing held out & .981 $\pm$ .000 & .969 $\pm$ .000 & .922 $\pm$ .004 & .880 $\pm$ .010\\

\hline
\hline

 \textbf{Short Summaries} &  &  & & \\
\hline

All questions & .483 $\pm$ .002 & .370 $\pm$ .049 & .412 $\pm$ .005 & .314 $\pm$ .100\\
Temporal questions & .465 $\pm$ .003 & .378 $\pm$ .040 & .421 $\pm$ .006 & .347 $\pm$ .091\\
All other (i.e. non-temporal) questions & .404 $\pm$ .003 & .276 $\pm$ .041 & .365 $\pm$ .005 & .227 $\pm$ .052\\
\hline
No Q\&A training on held out objects & .401 $\pm$ .003 & .277 $\pm$ .045 & .427 $\pm$ .003 & .288 $\pm$ .049\\
\hline
All objects and questions - nothing held out  & .571 $\pm$ .000 & .556 $\pm$ .006 & .517 $\pm$ .004 & .504 $\pm$ .022\\

\end{tabular}
\caption{\label{tab:transfer_more_metrics}
\textsc{rouge} and \textsc{bleu} scores of zero-shot long summarization when transferring representations learned from question answering to producing long (at the top) and short (at the bottom) summaries, broken down by question type used to learn the objects held out from summarization training. Results shown for episodes containing held-out objects in the validation sets in unseen and seen environments. The bottom two rows of the long and short summary sections show baselines with no question answering training on the held-out objects --- and therefore no transfer --- and a comparison to the fully trained model with nothing held out. This table complements Table 3 in the main body of the paper, which provides binary all-or-nothing accuracy scores for long summaries.}
\end{table*}

\section{Additional comparison metrics for zero-shot transfer}
\label{sec:app_zero_shot_transfer}
In Table \ref{tab:transfer_more_metrics} we provide additional metrics to understand the performance of zero-shot transfer from question answering to summarization. Short summaries tend to have lower \textsc{bleu} and \textsc{rouge} scores than the long summaries because the long summaries use a standardized set of words to describe actions and objects while the short summaries use a more diverse set of words and provide varying levels of detail.

\begin{table}
\begin{tabular}{lcc}
\hline
\textbf{Question type}   & \textbf{Val seen} & \textbf{Val unseen} \\
\hline

Ordinary Qs & .83 & .86\\
Extraordinary Qs & .64 & .76\\
\end{tabular}
\caption{\label{tab:ood}
Accuracy of one trained model on small test sets of out of distribution negative questions, by question and environment type.}
\end{table}

\section{Out of distribution negative questions}
Our model was trained on a set of questions which involved a certain set of actions and objects in a particular (household) environment. What would the model do if asked questions about actions it did not engage in with objects it had not seen, either at test time or during training? This would be a particular issue for yes/no questions if an agent were asked if it had engaged in an action it did not engage in and was not familiar with, as it may be expected to respond essentially randomly to such unfamiliar questions. 

To begin to investigate how such a model deals with these out of distribution negative questions, we developed two small test sets of questions:

1. Ordinary Out of Distribution Questions, consisting of questions about actions a household robot may be expected to take but which are not present in our dataset in any form, such as:

``did you clean the attic?''

``did you move the toys?''

``did you do the laundry?''

``did you water the plants?''

``did you take out the garbage?''

2. Extraordinary Out of Distribution Questions, consisting of questions totally unrelated to robots or household chores, such as:

''did you swim to the coral reef?''

``did you learn German?''

``did you fall in love?''

``did you kayak in the fjord?''

``did you graduate from college?''

Both test sets consist of fifty such questions (released along with our larger set of questions and answers paired to the \textsc{alfred} dataset). The correct answer to all of the questions was ``no''. The questions were run through a fully trained model (no held out objects). Results are shown in Table \ref{tab:ood}.

The model generally demonstrates a bias for answering ``no'' to out of distribution questions. Perhaps surprisingly, the model is much more likely to correctly answer ``no'' to \textit{Ordinary} Out of Distribution Questions than it is to \textit{Extraordinary} Out of Distribution Questions. 

These results, though on a small test set, suggest that the model has learned a bias toward answering, ``yes'', only when there is evidence in the input that an answer should be answered affirmatively. This is, of course, what a user would want. Further investigation of the circumstances under which it correctly answers out of distribution questions is warranted, as well as ways to improve the performance on out of distribution questions, especially unusual ones.




\end{appendices}


\bibliography{main.bib}


\end{document}